\documentclass[10pt,twocolumn]{article}

\input{glyphtounicode}
\usepackage[margin=0.75in,top=0.75in,bottom=0.75in]{geometry}
\usepackage[T1]{fontenc}
\usepackage{amsmath}
\usepackage{newtxtext,newtxmath}
\usepackage{dsfont}
\usepackage{booktabs}
\usepackage{graphicx}
\usepackage{xcolor}
\usepackage{microtype}
\usepackage{enumitem}
\usepackage{caption}
\usepackage{tabularx}
\usepackage{multirow}

\usepackage[colorlinks=true,linkcolor=blue!70!black,citecolor=blue!70!black,urlcolor=blue!70!black]{hyperref}

\setlist{nosep,leftmargin=*}

\title{\LARGE \textbf{Language Models Steer on Injected Uncertainty Tokens,\\but Fail to Self-Monitor In Situ}}

\author{
  \textbf{Daniel Theron} \\
}

\date{October 2026}

\begin{document}
\raggedbottom

\twocolumn[
  \begin{@twocolumnfalse}
    \maketitle
    \begin{abstract}
    An increasingly popular approach for triggering backtracking and error correction in step-by-step agentic reasoning involves prompting language models to verbalize their stepwise numerical confidence during generation. This strategy presupposes two capabilities: (1)~in-situ monitoring fidelity (does verbalized confidence drop when an error occurs?) and (2)~downstream steering responsiveness / control (does an uncertainty token causally steer the model to repair its error?). We evaluated four open-weights models (Qwen2.5-7B, Llama-3.1-8B, Mistral-7B, and Phi-4, 14.7B) across 15,200 attempts at solving math benchmark problems, isolating 2,017 reasoning errors on problems within each model's capability frontier. Across these failure points we applied 27,568 counterfactual continuations, and demonstrate a consistent decoupling between monitoring and control.
    On monitoring, all four models fail to detect their own errors. Verbalized confidence does not distinguish mistake steps from sound reasoning, yielding near-chance within-position AUROCs between 0.500 and 0.561. Even when a model makes its first reasoning error, its reported trajectory confidence remains flatly saturated at 95.1\% to 99.9\%, with noticeable drops ($\ge 10$ points) appearing on fewer than 9\% of errors in Llama and under 1.6\% in the other three models. In contrast, models respond readily to external uncertainty: injecting a low-confidence tag (10\%) at the error step significantly increases downstream correction across all models (+3.7 to +22.5 points over format-matched 95\% certainty; all $p \le 0.0022$), reaching up to 38.5\% under active rethinking instructions. However, injecting doubt onto already-correct steps causes models to discard valid reasoning, reducing solve rates by up to 22.8 points. We conclude that inference-time self-correction is bottlenecked by in-situ error detection rather than downstream repair capacity: models can readily steer toward recovery when prompted with doubt, but fail to self-monitor while generating their solutions.
    \end{abstract}
  \end{@twocolumnfalse}
]

\section{Introduction}

Iterative reasoning frameworks, tree-search decoders, and self-evaluating agents frequently rely on intermediate step assessments and verbalized evaluations to guide inference-time search, pruning, and backtracking \cite{kadavath2022language,shinn2023reflexion,yao2023tree}. In sequential Chain-of-Thought (CoT) reasoning, scaffolds increasingly prompt models to verbalize explicit confidence estimates after each step. These markers (ranging from local assessments $c_k$ to cumulative trajectory indicators $g_k$) are meant to let inference-time controllers decide whether to continue, backtrack, or prune. Relying on in-situ verbalized confidence for autonomous control motivates three core research questions:
\begin{itemize}
    \item \textbf{RQ1 (In-Situ Monitoring Fidelity):} Does verbalized confidence drop when a reasoning error occurs ($k^*$)? Can step confidence ($c_k$) or cumulative progress ($g_k$) discriminate valid derivation steps from erroneous ones?
    \item \textbf{RQ2 (Post-Dip Dynamics and the Confidence Ratchet):} Following a confidence dip, does the model exhibit an accelerated, unearned overconfidence rebound (a ``confidence ratchet'') that systematically masks the error in subsequent steps?
    \item \textbf{RQ3 (Counterfactual Causal Steering and False-Positive Costs):} Can externally injected uncertainty tokens at the failure boundary causally steer downstream recovery, and what collateral disruption does injection cause on valid reasoning?
\end{itemize}

While prompting models to verbalize confidence can capture coarse self-knowledge or improve upon degraded conditional logit probabilities \cite{kadavath2022language,tian2023just}, zero-shot verbalized confidence frequently suffers from severe overconfidence and poor calibration in aggregate \cite{lin2022teaching,xiong2024can}. In parallel, studies on intrinsic self-correction show that language models struggle to correct their reasoning without external feedback \cite{huang2024large}, though Tyen et~al.~\cite{tyen2024llms} demonstrated that models can repair errors if provided with oracle step locations. However, prior work has not disentangled \textbf{in-situ monitoring failure} from \textbf{downstream steering unresponsiveness}: when an LLM fails to self-correct during generation, is it because structured verbalization fails to register errors in situ, or because the model cannot execute downstream repairs once uncertainty is established in context? Does this decoupling reflect an autoregressive commitment bias, an evaluative incapacity during generation, or a bottleneck of the verbalization format?

We investigate in-situ monitoring and causal steering across up to 550 mathematical reasoning problems from GSM8K \cite{cobbe2021training} and the MATH-500 test subset \cite{hendrycks2021measuring,lightman2023let} across four open-weights instruction-tuned models spanning diverse developer lineages and scales: Alibaba (Qwen2.5-7B-Instruct), Meta (Llama-3.1-8B-Instruct), Mistral AI (Mistral-7B-Instruct-v0.3), and Microsoft (Phi-4, 14.7B). Using \texttt{Claude Opus 5.5} as an automated step-level process judge (validated under the deployment condition with full step blocks and reference solutions), we evaluate:
\begin{enumerate}
    \item \textbf{In-Situ Monitoring \& Ratchet Dynamics (RQ1 \& RQ2; Phase 1):} From 15,200 primary reasoning traces ($4{,}400$ each for Qwen and Llama; $3{,}200$ each for Mistral and Phi-4), format validation and contested-item filtering yield 3,873 valid traces across contested problems (containing 2,017 verified error traces). Across these, we trace structured verbalized step confidence ($c_k$) and trajectory progress ($g_k$) during active in-situ generation, testing whether confidence drops at the first error ($k^*$), evaluating step-level discrimination AUROC, and testing whether confidence rebounds abnormally following dips.
    \item \textbf{Counterfactual Causal Steering (RQ3; Phase 2):} On contested items (problems where the model succeeded in $\ge 1$ and failed in $\ge 1$ of $n_{\text{samp}} = 8$ primary traces), we externally manipulate the confidence tag at $k^*$ across five counterfactual conditions (\texttt{Native}, \texttt{High}, \texttt{Mid}, \texttt{Low}, and a verbal \texttt{Positive Control}) and two prompt arms (\texttt{Passive} vs.\ \texttt{Active}), generating 27,568 continuations on error and matched clean prefixes.
\end{enumerate}

In all, these observations indicate that structured verbalized confidence fails in agentic workflows not because models cannot condition on uncertainty tokens (which reliably steer downstream repair), but because active in-situ generation fails to detect errors when they occur.

\section{Related Work}

\textbf{Verbalized Confidence and Model Calibration.}
Calibration of neural representations traditionally evaluates whether predicted class probabilities match empirical accuracy \cite{guo2017calibration}. In autoregressive language models, Kadavath et~al.~\cite{kadavath2022language} evaluated verbalized self-evaluation, demonstrating that models can express coarse calibration on question answering. Lin et~al.~\cite{lin2022teaching} taught models to express uncertainty in words, while Tian et~al.~\cite{tian2023just} established that explicitly prompting models for verbalized probabilities improves upon degraded conditional logit probabilities. However, Xiong et~al.~\cite{xiong2024can} found that across multi-step mathematical reasoning, verbalized confidence suffers from extreme overconfidence and poor calibration. Our work extends these findings by analyzing confidence at the step level during active generation, explicitly contrasting local step evaluation ($c_k$) against cumulative trajectory viability ($g_k$).

\textbf{Process Reward Models and Step Verification.}
Rather than evaluating complete derivations with outcome reward models (ORMs; \cite{cobbe2021training}), Process Reward Models (PRMs; \cite{lightman2023let}) evaluate intermediate reasoning steps. Wang et~al.~\cite{wang2024mathshepherd} introduced Math-Shepherd to automate step-level verifier supervision via Monte Carlo tree rollouts. Zheng et~al.~\cite{zheng2024processbench} introduced ProcessBench to systematically benchmark automated process verifiers against human step annotations. While PRM research primarily focuses on training external verifiers to score candidate reasoning paths, our work addresses whether generating models can self-monitor in situ, and quantifies the operational precision required for external PRMs to steer generation effectively.

\textbf{Intrinsic Self-Correction and Error Localization.}
Studies on intrinsic self-correction demonstrate that language models struggle to correct their reasoning without external feedback \cite{huang2024large}, often suffering from performance degradation when prompted to re-evaluate their answers. In a critical survey, Kamoi et~al.~\cite{kamoi2024when} concluded that self-correction succeeds mainly when reliable external feedback is available, pointing to feedback generation (i.e., error detection) as the key bottleneck. Tyen et~al.~\cite{tyen2024llms} demonstrated that while models cannot locate their own reasoning errors, they can successfully correct them if provided with the exact error step location. To reduce performance degradation, Kumar et~al.~\cite{kumar2024score} introduced SCoRe, using multi-turn reinforcement learning to cultivate genuine self-correction. Our work clarifies this distinction by formally disentangling in-situ monitoring failure from downstream steering insensitivity.

\textbf{Test-Time Scaling and Reasoning Scaffolds.}
Inference-time search algorithms (including Tree-of-Thought \cite{yao2023tree}, Reflexion \cite{shinn2023reflexion}, and test-time compute scaling \cite{snell2024scaling}) depend on step-level assessments to guide backtracking and branch pruning. Muennighoff et~al.~\cite{muennighoff2025s1} showed that appending simple test-time rethink prompts (e.g., ``Wait...'') yields effective test-time scaling. Concurrently, native reasoning models (e.g., OpenAI o1 \cite{openai2024o1}, DeepSeek-R1 \cite{deepseek2025r1}) incorporate extensive internal exploration and backtracking developed through reinforcement learning. Our counterfactual steering analysis provides an empirical foundation for test-time intervention in standard instruction-tuned models, establishing break-even precision thresholds for external verification monitors in deployment.

\section{Experimental Framework}

\subsection{Elicitation Protocol and Formulations}
To elicit step-by-step reasoning and verbalized confidence without conflating local correctness with trajectory-level viability, models are prompted with structured formatting directives:
\begin{quote}
\small
\textbf{Step $i$:} $\langle\text{reasoning}\rangle$\\
\texttt{[Step confidence: $X$\%]}\\
\texttt{[On-track: $Y$\%]}
\end{quote}
Formally, for a derivation of $K$ steps, let:
\begin{itemize}
    \item $c_k \in [0, 100]$ denote the verbalized local confidence in step $k$.
    \item $g_k \in [0, 100]$ denote the verbalized cumulative (on-track) confidence that all steps so far, including step $k$, are correct.
    \item $k^* = \min \{ k \in \{1, \dots, K\} : \text{step } k \text{ is logically invalid} \}$ denote the index of the earliest reasoning error step (with $k^* = -1$ denoting clean, error-free derivations).
\end{itemize}
While prompting for both $c_k$ and $g_k$ enables evaluating whether models distinguish local step correctness from cumulative trajectory viability during primary generation (Phase 1), our Phase 2 steering interventions deliberately set both tags concurrently to congruent values (e.g., 95\%/95\% or 10\%/10\%) to deliver an unambiguous uncertainty signal and prevent confounding from contradictory token cues (Appendix~\ref{app:tags}). Downstream steering thus evaluates the joint causal effect of step and trajectory doubt, while isolating fine-grained permutations ($c \neq g$) remains an open direction.

\subsection{Models and Datasets}
We evaluate four widely-used open-weights instruction-tuned models across diverse architectures and parameter scales:
\begin{itemize}
    \item \textbf{Qwen2.5-7B-Instruct} (abbreviated \textit{Qwen})
    \item \textbf{Llama-3.1-8B-Instruct} (abbreviated \textit{Llama})
    \item \textbf{Mistral-7B-Instruct-v0.3} (abbreviated \textit{Mistral})
    \item \textbf{Phi-4} (14.7B; abbreviated \textit{Phi-4})
\end{itemize}
Inference is executed via vLLM (\texttt{max\_model\_len = 4096}) with nucleus sampling ($T = 0.7$, $\text{top-}p = 0.95$, $n_{\text{samp}} = 8$, new token cap $\text{max\_tokens} = 1{,}536$). The evaluation problems span GSM8K test items \cite{cobbe2021training} and the MATH-500 test subset \cite{hendrycks2021measuring,lightman2023let} spanning intermediate difficulty (Levels 2--4) across four subjects (algebra, prealgebra, number theory, and counting \& probability; excluding Level 1 trivial arithmetic, Level 5 Olympiad outliers, and geometry/precalculus requiring spatial diagram interpretation). For Qwen and Llama, generations span 550 problems (400 GSM8K, 150 MATH-500); for Mistral and Phi-4, generations span 400 problems (250 GSM8K, 150 MATH-500), generating $4{,}400 \times 2 + 3{,}200 \times 2 = 15{,}200$ primary traces. All final answers are graded with symbolic equivalence verification using \texttt{math-verify}.

\begin{figure*}[t]
\centering
\includegraphics[width=0.84\textwidth]{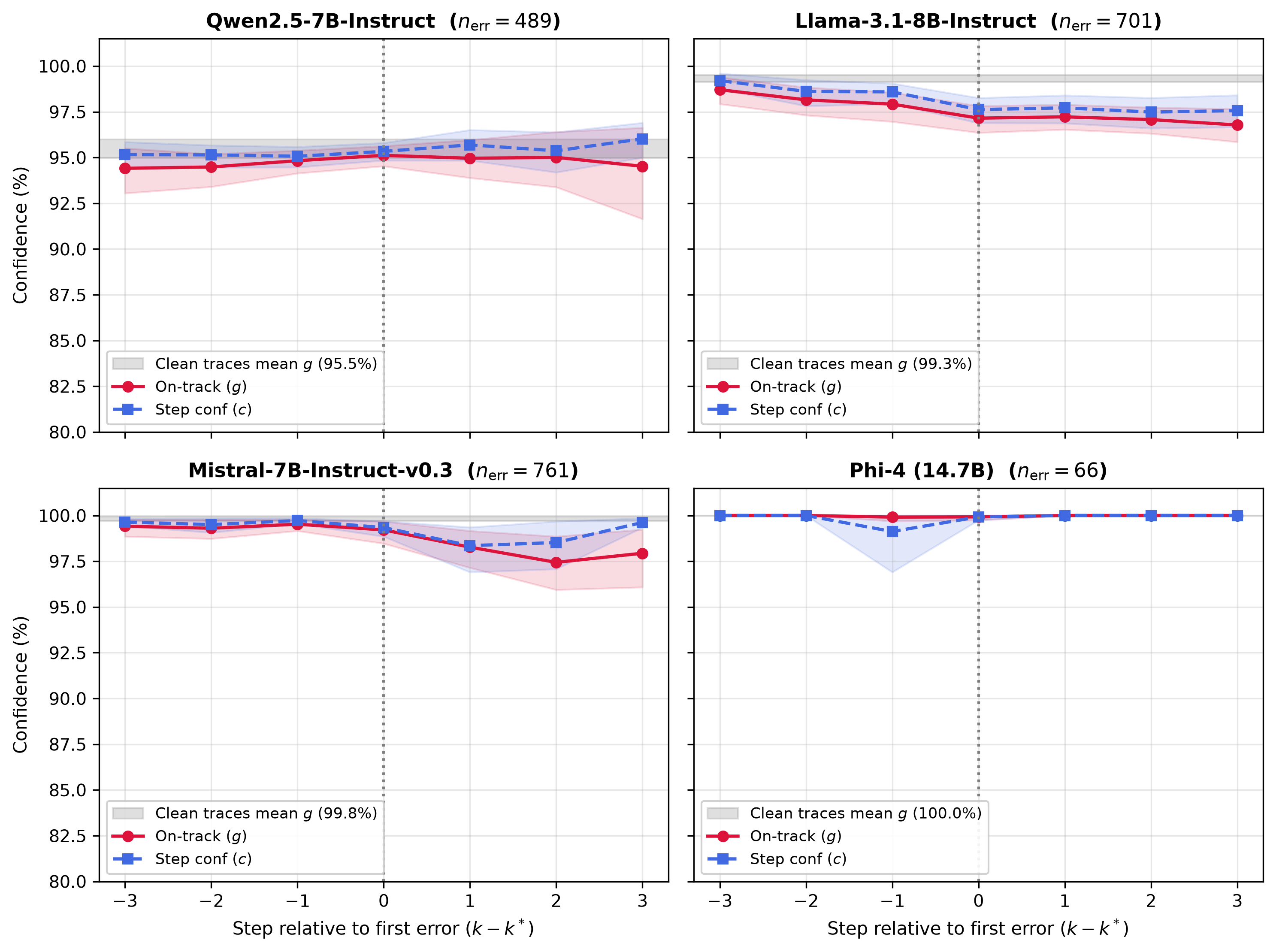}
\caption{\textbf{Confidence Dynamics at the First Error Step ($k^*$).} Mean on-track confidence ($g_k$, red circle) and step confidence ($c_k$, blue square) centered at the earliest error step ($k - k^* = 0$) with 95\% bootstrap confidence intervals across four models: \texttt{Qwen2.5-7B-Instruct}, \texttt{Llama-3.1-8B-Instruct}, \texttt{Mistral-7B-Instruct-v0.3}, and \texttt{Phi-4} (14.7B). Sample sizes at relative steps $k - k^* < 0$ exclude step-1 errors ($k^* = 1$, leaving $n = 429$ Qwen, $551$ Llama, $646$ Mistral, and $57$ Phi-4 at $k - k^* = -1$). Trajectories remain flatly saturated at 95--100\% across reasoning errors.}
\label{fig:fig1}
\end{figure*}

\subsection{Process Verification with Claude Opus 5.5}
To localize $k^*$ across thousands of derivations, we employ \texttt{claude-opus-5-5} via the Anthropic API under an optimized deployment protocol:
\begin{itemize}
    \item The judge is supplied with the problem statement, gold reference derivation, and gold final answer.
    \item Student steps are presented in complete blocks with all confidence tags stripped, ensuring the judge is blinded to student self-evaluations.
\end{itemize}

\textbf{Validation Gate and Audit:}
As documented in Table~\ref{tab:audit}, the judge protocol underwent two empirical validations:
\begin{enumerate}
    \item \textbf{ProcessBench Validation:} On $N=200$ items from ProcessBench \cite{zheng2024processbench} (100 from the \texttt{gsm8k} split, 100 from \texttt{math}, random seed 0; 100 error traces, 100 clean traces), the judge achieved \textbf{85.0\% exact error step accuracy} (85/100 error traces localized exactly, 94.0\% within $\pm 1$ step) and \textbf{97.0\% clean accuracy} (97/100 correctly identifying no error), passing our planned $\ge 60\%$ validity threshold.
    \item \textbf{Blind Human Hand-Audit:} An author blind hand-check of 30 experimental traces drawn from contested items (15 Qwen, 15 Llama, random seed 42, sampled from incorrect-answer valid traces and evaluated blind to judge labels) yielded \textbf{93.3\% exact agreement} (28/30; Wilson 95\% CI: [78.7\%, 98.2\%]) and \textbf{96.7\% within $\pm 1$ agreement} (29/30; Wilson 95\% CI: [83.3\%, 99.4\%]).
\end{enumerate}
From 15,200 primary generations, format validation (verifying 1-indexed consecutive numbering, both confidence tags, and a boxed final answer) passed 3,656 traces in Qwen (83.1\%), 4,110 traces in Llama (93.4\%), 2,723 traces in Mistral (85.1\%), and 2,830 traces in Phi-4 (88.4\%). Contested items were identified strictly on format-valid traces requiring $\ge 4$ valid samples and intermediate solve accuracy ($15\% \le \text{accuracy} \le 85\%$) via \texttt{math-verify}. This yielded 1,168 valid traces across $I=177$ contested items in Qwen (mean 6.60 valid samples/item), 1,326 valid traces across $I=184$ contested items in Llama (mean 7.21 valid samples/item), 1,209 valid traces across $I=177$ contested items in Mistral (mean 6.83 valid samples/item), and 170 valid traces across $I=25$ contested items in Phi-4 (mean 6.80 valid samples/item), totaling 3,873 valid contested traces. In Phi-4, higher baseline accuracy on GSM8K and Levels 2--4 of MATH-500 resulted in fewer items falling within the intermediate accuracy window, yielding 25 contested items that isolate problems at the model's performance frontier. The process judge localized all 3,873 traces with zero parsing failures. Cross-classifying final-answer correctness against process localization identifies clean traces ($k^* = -1$) comprising 675 traces in Qwen, 613 in Llama, 436 in Mistral, and 97 in Phi-4 (all correct), while error traces ($k^* \ge 1$) comprise 489 in Qwen, 701 in Llama, 761 in Mistral, and 66 in Phi-4 (including 34 Qwen, 120 Llama, 72 Mistral, and 5 Phi-4 traces that reached correct answers despite earlier errors). Traces where \texttt{math-verify} graded the answer incorrect but the judge detected no step error occurred in only 4 Qwen (0.34\%), 11 Llama (0.83\%), 10 Mistral (0.83\%), and 7 Phi-4 (4.1\%) traces, alongside 1 Llama trace and 2 Mistral traces with $k^*$ exceeding trace length; these 35 rare discordant traces were excluded from analysis to preserve ground-truth integrity (Appendix~\ref{app:judge}). Restricting counterfactual analysis to contested items ensures that failure to recover reflects trajectory steering difficulties rather than problems beyond the model's achievable frontier.

\begin{table}[b]
\centering
\small
\setlength{\tabcolsep}{4pt}
\caption{\textbf{Process Judge Validation and Blind Hand-Audit.} Accuracy on ProcessBench ($N=200$, 100 error items) and blind human audit on contested experimental traces ($N=30$).}
\label{tab:audit}
\begin{tabular}{@{}lcc@{}}
\toprule
\textbf{Evaluation Benchmark} & \textbf{Exact Step} & \textbf{Within $\pm 1$ Step} \\
\midrule
ProcessBench Gate ($\ge 60\%$) & \textbf{85.0\%} (85/100) & \textbf{94.0\%} (94/100) \\
Blind Human Hand-Audit & \textbf{93.3\%} (28/30) & \textbf{96.7\%} (29/30) \\
\bottomrule
\end{tabular}
\end{table}

\begin{table*}[t]
\centering
\footnotesize
\setlength{\tabcolsep}{1.0pt}
\renewcommand{\arraystretch}{0.90}
\caption{\textbf{In-Situ Monitoring Metrics at Failure Boundary ($k^*$).} Evaluated on contested items across three nested sample tiers across four models (\texttt{Qwen2.5-7B-Instruct}, \texttt{Llama-3.1-8B-Instruct}, \texttt{Mistral-7B-Instruct-v0.3}, and \texttt{Phi-4}, 14.7B): Tier 1: all verified error traces ($n=489$ Qwen, $701$ Llama, $761$ Mistral, $66$ Phi-4; $c_{k^*}, g_{k^*}$); Tier 2: boundary-evaluable traces with $k^* \ge 2$ ($n=429$ Qwen, $551$ Llama, $646$ Mistral, $57$ Phi-4; $D_{\text{err}}$, dip rates); Tier 3: within-item step-matched pairs ($n=379$ Qwen, $451$ Llama, $458$ Mistral, $44$ Phi-4; $\Delta D = D_{\text{err}} - D_{\text{ctl}}$, where Tier 3 error means are $D_{\text{err}} = -0.47, +1.38, +0.58, 0.00$ pts and clean control means are $D_{\text{ctl}} = -0.83, +0.15, +0.14, 0.00$ pts). Clean-trace Mean $g$ is averaged across all steps of clean traces on contested items ($n=675$ Qwen, $613$ Llama, $436$ Mistral, $97$ Phi-4). Control dip rates evaluate matched clean transitions at depth $k^*$. Coherence evaluates Tier 1. Step and trace discrimination is pooled across all steps in each model; steps $k \ge k^*$ are labeled off-track (errors) and $k < k^*$ as on-track. Within-position AUROC evaluates discrimination within each step index $k$, weighted by stratum sample size $n_k$, stripping out mechanical step-depth confounding. Brackets report 95\% item-clustered bootstrap confidence intervals (10,000 resamples). $^*$Two-sided Wilcoxon signed-rank test on unclustered matched pairs ($p < 0.05$; item-aggregated differences yield $p = 0.0591$ for Qwen, $p = 0.0001$ for Llama, $p = 0.1031$ for Mistral, $p = 1.0000$ for Phi-4). $^\dagger$RQ2 evaluates subsequent-step recovery $R_d = (g_{k+1}-g_k)/(g_{k-1}-g_k)$ across non-terminal dips ($2 \le k < K$).}
\label{tab:monitoring}
\begin{tabular*}{\textwidth}{@{\extracolsep{\fill}}lcccc}
\toprule
\textbf{Metric} & \textbf{Qwen2.5-7B} & \textbf{Llama-3.1-8B} & \textbf{Mistral-7B} & \textbf{Phi-4} \\
\midrule
\multicolumn{5}{l}{\textit{Tier 1: All Verified Error Traces ($n=489$ Qwen, $n=701$ Llama, $n=761$ Mistral, $n=66$ Phi-4)}} \\
Step Confidence at $k^*$ ($c_{k^*}$) & 95.3\% [94.9, 95.8] & 97.6\% [96.9, 98.3] & 99.3\% [98.9, 99.7] & 99.9\% [99.7, 100.0] \\
On-track Confidence at $k^*$ ($g_{k^*}$) & 95.1\% [94.6, 95.7] & 97.2\% [96.3, 98.0] & 99.2\% [98.5, 99.7] & 99.9\% [99.7, 100.0] \\
Clean-trace Mean $g$ & 95.5\% & 99.3\% & 99.8\% & 100.0\% \\
\midrule
\multicolumn{5}{l}{\textit{Tier 2: Boundary-Evaluable Error Traces ($k^* \ge 2$; $n=429$ Qwen, $n=551$ Llama, $n=646$ Mistral, $n=57$ Phi-4)}} \\
Boundary Drop ($D_{\text{err}} = g_{k^*-1} - g_{k^*}$) & $-0.52$ pts $[-0.81, -0.24]$ & $+1.01$ pts $[0.29, 1.74]$ & $+0.35$ pts $[-0.20, 1.15]$ & $0.00$ pts $[0.00, 0.00]$ \\
Dip Rate ($D_{\text{err}} \ge 10$ pts at $k^*$): Error vs.\ Control & \textbf{1.6\%} (7/429) vs.\ 1.4\% & \textbf{8.7\%} (48/551) vs.\ 4.1\% & \textbf{0.8\%} (5/646) vs.\ 0.3\% & \textbf{0.0\%} (0/57) vs.\ 0.0\% \\
\midrule
\multicolumn{5}{l}{\textit{Tier 3: Within-Item Step-Matched Pairs ($n=379$ Qwen, $n=451$ Llama, $n=458$ Mistral, $n=44$ Phi-4)}} \\
Paired Contrast ($\Delta D = D_{\text{err}} - D_{\text{ctl}}$) & $+0.36$ pts $[-0.10, 0.84]$ & $+1.23$ pts $[0.62, 1.94]$ & $+0.45$ pts $[-0.26, 1.51]$ & $0.00$ pts $[0.00, 0.00]$ \\
\quad Wilcoxon signed-rank test & $p = 0.0118^*$ & $p = 0.0001^*$ & $p = 0.9140$ & $p = 1.0000$ \\
\midrule
\multicolumn{5}{l}{\textit{Coherence at Error Boundary ($k^*$)}} \\
Local vs.\ Trajectory Gap ($c_{k^*} - g_{k^*}$) & $+0.21$ pts $[-0.01, 0.47]$ & $+0.47$ pts $[-0.19, 1.19]$ & $+0.13$ pts $[-0.13, 0.48]$ & $0.00$ pts $[0.00, 0.00]$ \\
Compounding Deficit ($g_{k^*} - 100 \prod_{i=1}^{k^*} (c_i / 100)$) & $+9.20$ pts $[8.02, 10.31]$ & $+1.68$ pts $[0.73, 2.66]$ & $+0.63$ pts $[0.15, 1.16]$ & $+0.72$ pts $[0.00, 2.74]$ \\
\midrule
\multicolumn{5}{l}{\textit{Step and Trace Discrimination (Pooled Across Steps)}} \\
Step AUROC ($100 - g_k$, error vs.\ valid) & \textbf{0.508} [0.478, 0.539] & \textbf{0.600} [0.580, 0.620] & \textbf{0.526} [0.513, 0.539] & \textbf{0.502} [0.500, 0.507] \\
Step AUROC ($100 - c_k$, error vs.\ valid) & \textbf{0.510} [0.480, 0.539] & \textbf{0.580} [0.561, 0.598] & \textbf{0.521} [0.510, 0.534] & \textbf{0.502} [0.500, 0.507] \\
Position-only Baseline AUROC ($k$) & 0.692 [0.666, 0.721] & 0.708 [0.684, 0.732] & 0.739 [0.715, 0.762] & 0.706 [0.643, 0.780] \\
Within-position AUROC ($100 - g_k \mid k$) & 0.537 [0.500, 0.576] & 0.561 [0.544, 0.579] & 0.525 [0.512, 0.539] & \textbf{0.500} [0.500, 0.502] \\
Within-position AUROC ($100 - c_k \mid k$) & 0.536 [0.500, 0.575] & 0.552 [0.535, 0.570] & 0.522 [0.510, 0.535] & \textbf{0.500} [0.500, 0.502] \\
Trace AUROC ($\min g \to \text{wrong}$) & 0.523 [0.485, 0.558] & 0.550 [0.523, 0.577] & 0.527 [0.512, 0.541] & 0.508 [0.500, 0.528] \\
\midrule
\multicolumn{5}{l}{\textit{RQ2: Post-Dip Recovery Regression on Dips ($n=117$ Pooled Dips)}} \\
RQ2 Dips ($n_{\text{err}} / n_{\text{clean}}$)$^\dagger$ & 6 / 19 (exploratory) & 44 / 44 (regression sample) & 3 / 1 (exploratory) & 0 / 0 (exploratory) \\
RQ2 Recovery Slope ($\beta_R$) & $\beta_R = -0.290$ $[-0.99, 0.41]$ & $\beta_R = -0.075$ $[-0.27, 0.12]$ & --- & --- \\
\quad $p$-value & $p = 0.4169$ & $p = 0.4462$ & --- & --- \\
Pooled RQ2 Effect ($\beta_R$) & \multicolumn{4}{c}{$\beta_R = -0.071$ $[-0.26, 0.12]$, $p = 0.471$ ($n = 117$; model FE: $\beta_R = -0.105$, $p = 0.288$)} \\
\bottomrule
\end{tabular*}
\end{table*}

\section{RQ1: In-Situ Monitoring Fidelity}

\subsection{Flat Saturation and the Sample Funnel}
Figure~\ref{fig:fig1} and Table~\ref{tab:monitoring} present in-situ monitoring trajectories centered at the earliest error step ($k - k^* = 0$). Verbalized confidence exhibits consistent \textbf{flat saturation} across all four models:
\begin{itemize}
    \item \textbf{Sample Filtering Funnel:} Traces are analyzed across three nested tiers based on data requirements:
    \textbf{(i)~\textit{All Verified Errors}} ($n=489$ Qwen, $701$ Llama, $761$ Mistral, $66$ Phi-4): all contested error traces with verified $k^*$, used for event-study trajectories in Figure~\ref{fig:fig1} and boundary confidences ($c_{k^*}, g_{k^*}$);
    \textbf{(ii)~\textit{Boundary Drop Traces}} ($k^* \ge 2$; $n=429$ Qwen, $551$ Llama, $646$ Mistral, $57$ Phi-4): traces with an antecedent step $k^*-1$ required to evaluate boundary drop $D_{\text{err}} = g_{k^{*}-1} - g_{k^{*}}$ (excluding immediate step-1 blunders); and
    \textbf{(iii)~\textit{Step-Matched Pairs}} ($n=379$ Qwen, $451$ Llama, $458$ Mistral, $44$ Phi-4): traces paired with a clean transition at the exact same step depth on the \textit{same problem item} to compute $\Delta D = D_{\text{err}} - D_{\text{ctl}}$ (averaging clean control transitions for that item).
    All bootstrap confidence intervals cluster by problem item ($I=177$ Qwen, $184$ Llama, $177$ Mistral, $25$ Phi-4; 10,000 resamples). The two-sided Wilcoxon signed-rank test evaluates paired transitions across all matched traces ($n_{\text{paired}}$); aggregating differences by problem item yields consistent rank patterns ($p = 0.0591$ for Qwen, $p = 0.0001$ for Llama, $p = 0.1031$ for Mistral, and $p = 1.0000$ for Phi-4).
    \item At the earliest reasoning error step $k^{*}$, mean step confidence $c_{k^*}$ is \textbf{95.3\%} for Qwen, \textbf{97.6\%} for Llama, \textbf{99.3\%} for Mistral, and \textbf{99.9\%} for Phi-4. Trajectory on-track confidence $g_{k^*}$ remains at \textbf{95.1\%} (Qwen), \textbf{97.2\%} (Llama), \textbf{99.2\%} (Mistral), and \textbf{99.9\%} (Phi-4).
    \item Raw boundary drops, $D_{\text{err}} = g_{k^{*}-1} - g_{k^{*}}$, are minimal: $-0.52$ pts in Qwen, $+1.01$ pts in Llama, $+0.35$ pts in Mistral, and $0.00$ pts in Phi-4.
    \item \textbf{Paired Control Analysis:} Within the matched Tier 3 sample, error transitions have mean $D_{\text{err}} = -0.47$ pts for Qwen ($+1.38$ Llama, $+0.58$ Mistral, $0.00$ Phi-4), while clean control transitions have mean $D_{\text{ctl}} = -0.83$ pts for Qwen ($+0.15$ Llama, $+0.14$ Mistral, $0.00$ Phi-4). In Qwen, the apparent sign difference between raw drops ($-0.52$ pts) and paired contrasts ($\Delta D = +0.36$ pts) is driven by clean control drift: clean traces slightly increase confidence at depth $k^*$ ($D_{\text{ctl}} = -0.83$ pts), so error traces drop relative to clean baselines despite both exhibiting negative raw drops. Testing paired differentials $\Delta D = D_{\text{err}} - D_{\text{ctl}}$ yields $+0.36$ pts for Qwen ($p = 0.0118$, Holm-adjusted $p = 0.0235$), $+1.23$ pts for Llama ($p = 0.0001$, Holm-adjusted $p = 0.0003$), $+0.45$ pts for Mistral ($p = 0.9140$, Holm-adjusted $p = 0.9140$), and $0.00$ pts for Phi-4 ($p = 1.0000$), where the Holm adjustment family spans the three primary research endpoints within each model ($M=3$: RQ1 $\Delta D$, RQ2 $\beta_R$, RQ3 Low vs.\ High). In Qwen, the divergence between the item-clustered bootstrap CI ($[-0.10, 0.84]$, spanning zero) and the trace-level Wilcoxon test ($p = 0.0118$) reflects both cluster-robust variance and the difference between the mean and rank median on distributions with point masses at zero. While Qwen and Llama show small positive rank shifts, point estimates remain near zero ($\le 1.23$ pts), far too subtle to support operational filtering.
    \item \textbf{Absence of Dips:} A perceptible confidence dip on cumulative on-track confidence ($D_{\text{err}} = g_{k^*-1} - g_{k^*} \ge 10\text{ pts}$ at $k^{*}$) occurs in only \textbf{1.6\%} (7/429) of Qwen, \textbf{8.7\%} (48/551) of Llama, \textbf{0.8\%} (5/646) of Mistral, and \textbf{0.0\%} (0/57) of Phi-4 error traces. Thus, over 91\% of error steps in Llama and over 98\% in the other three models pass without any perceptible drop in verbalized confidence.
\end{itemize}

\begin{figure*}[t]
\centering
\includegraphics[width=0.84\textwidth]{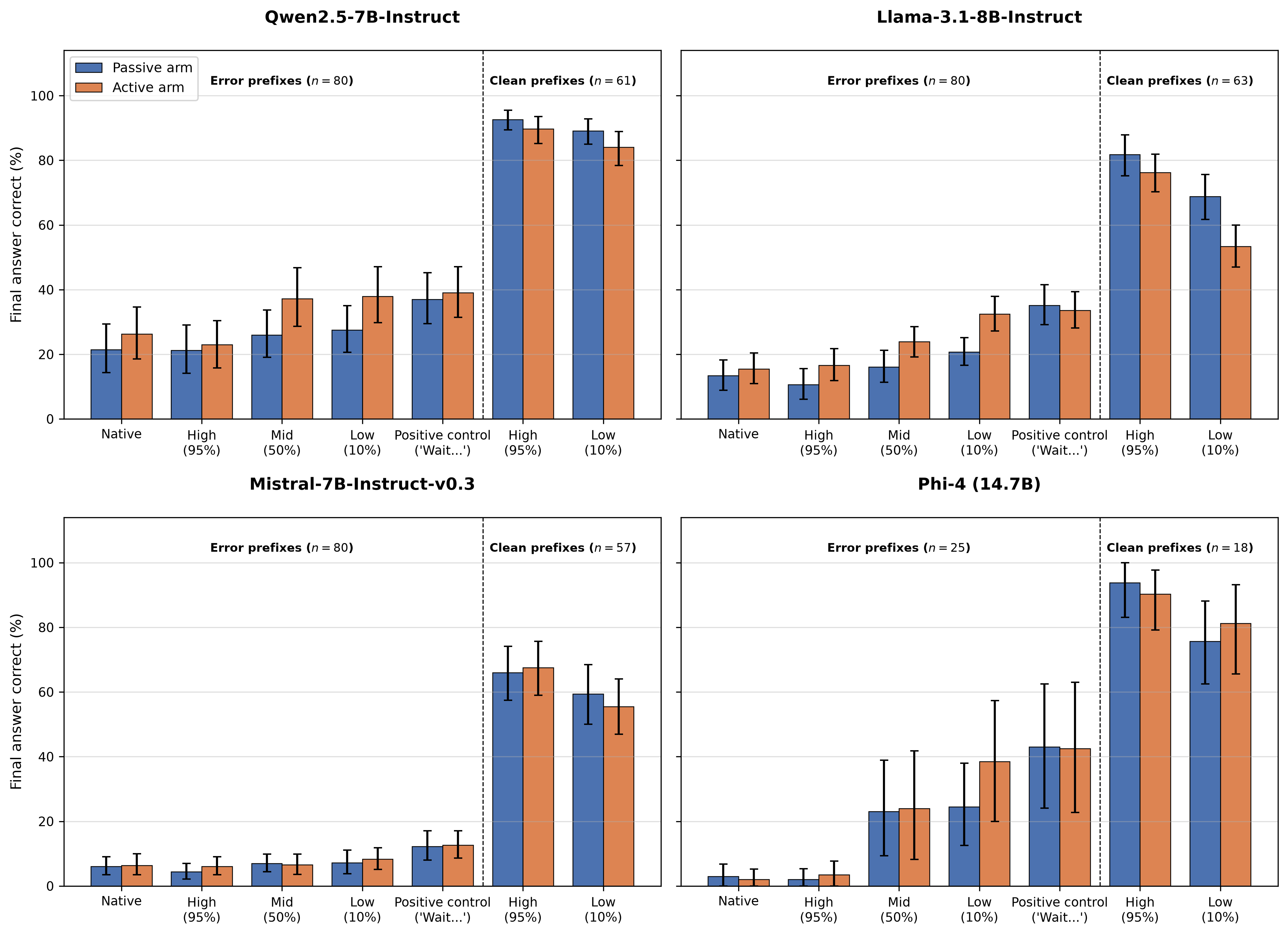}
\caption{\textbf{Counterfactual Causal Steering Across Conditions and Arms for Four Models.} Percentage of continuations achieving correct final answers after counterfactual tag injection at $k^*$, with 95\% cluster-robust confidence intervals across \texttt{Qwen2.5-7B-Instruct}, \texttt{Llama-3.1-8B-Instruct}, \texttt{Mistral-7B-Instruct-v0.3}, and \texttt{Phi-4} (14.7B). Evaluated across 27,568 continuations on error and clean prefixes. Injecting Low (10\%) confidence tags significantly increases correction rates over High (95\%) certainty tags across all four models ($p \le 0.0022$ in passive arms; $p \le 0.0341$ in active arms).}
\label{fig:fig2}
\end{figure*}

\subsection{Local-vs-Trajectory Gap and Failure to Compound Uncertainty}
In multi-step derivations where an unrecovered error compromises subsequent reasoning, cumulative trajectory viability $g_k$ is logically bounded by constituent step correctness. Under a marginal interpretation, where $c_i = P(\text{step } i \text{ valid})$, the upper Fréchet bound (monotonicity bound) requires $g_k \le \min_{i \le k} c_i \le c_k$. Under a conditional interpretation, where $c_i = P(\text{step } i \text{ valid} \mid \text{history valid})$, the probability chain rule yields $g_k = 100 \prod_{i=1}^k (c_i / 100)$ exactly.

In practice, all four models exhibit a local-versus-trajectory gap ($c_{k^*} - g_{k^*}$) near zero ($+0.21$ pts in Qwen, $+0.47$ pts in Llama, $+0.13$ pts in Mistral, and $0.00$ pts in Phi-4; Table~\ref{tab:monitoring}). Models do not accumulate uncertainty across steps: $g_k$ tracks local step confidence $c_k$ almost identically rather than decaying over time. Under the conditional chain-rule interpretation, $g_{k^*}$ exceeds $100 \prod_{i=1}^{k^*} (c_i / 100)$ across models ($+9.20$ pts in Qwen, $+1.68$ pts in Llama, $+0.63$ pts in Mistral, $+0.72$ pts in Phi-4). Rather than tracking accumulating trajectory risk, models treat $g_k$ as an uncalibrated repetition of local step confidence.

\vspace{0.3em}\noindent\textbf{Step and Trace Discrimination vs.\ Step-Depth Confounding.}
Across all pooled steps (labeling steps $k \ge k^*$ in contested traces as off-track and $k < k^*$ as on-track), step-level error discrimination via trajectory uncertainty ($100 - g_k$) yields an AUROC of \textbf{0.508} for Qwen, \textbf{0.600} for Llama, \textbf{0.526} for Mistral, and \textbf{0.502} for Phi-4 (Table~\ref{tab:monitoring}). While labeling $k \ge k^*$ as off-track aligns with cumulative trajectory invalidity ($g_k$), it conflates downstream algebraic deduction with error detection for local step confidence ($c_k$): steps following $k^*$ often execute valid algebraic deductions from an earlier erroneous value, so high local confidence $c_k$ on those steps is logically valid rather than a monitoring failure. Nonetheless, evaluating local step uncertainty ($100 - c_k$) yields similarly degraded AUROCs (0.510, 0.580, 0.521, and 0.502).

Because labeling all steps $k \ge k^*$ as off-track mechanically concentrates errors at larger step indices $k$, a naive baseline heuristic using solely the step index $k$ achieves AUROCs between 0.692 and 0.739 across models. Conditioning on step index via within-position AUROC (computing AUROC within each step index $k$ and weighting by stratum sample size $n_k$) removes this depth artifact, reducing discrimination to 0.537 in Qwen, 0.561 in Llama, 0.525 in Mistral, and \textbf{0.500} in Phi-4 for trajectory confidence $g_k$, and 0.536 (Qwen), 0.552 (Llama), 0.522 (Mistral), and \textbf{0.500} (Phi-4) for local step confidence $c_k$ (Table~\ref{tab:monitoring}). A sensitivity analysis excluding the 231 error traces that recovered downstream to correct answers yields identical near-chance within-position AUROCs ($\le 0.56$), confirming that label contamination does not drive the monitoring failure. In all, these results show that in-situ verbalized numerical confidence fails to discriminate valid reasoning steps from errors.

\section{RQ2: Testing the Confidence Ratchet}
The \textbf{Confidence Ratchet} hypothesis posits that if an LLM registers doubt at an error step ($g_{k-1} - g_k \ge 10$ pts), autoregressive continuation momentum drives an unearned overconfidence rebound in subsequent steps, ``ratcheting'' confidence back up while compounding the error. We evaluate the 1-step recovery ratio $R_d = (g_{k+1} - g_k)/(g_{k-1} - g_k)$ (winsorized to $[-1, 2]$) following any dip $d$ at step $k$, fitting an item-clustered OLS regression:
\begin{equation}
R_d = \alpha + \beta_R \mathds{1}[\text{error dip}_d] + \beta_{\text{size}} (g_{k-1} - g_k)_d + \beta_{\text{prev}} g_{k-1, d} + \epsilon_d
\label{eq:ratchet}
\end{equation}
where $\mathds{1}[\text{error dip}_d] = 1$ for dips at the earliest error step $k^*$, and $0$ for baseline dips in clean traces (controlling for regression to the mean). Baseline clean dips were exhaustively collected across non-terminal steps ($2 \le k < K$) in clean contested traces ($n_{\text{clean}} = 19$ in Qwen, $44$ in Llama, $1$ in Mistral, and $0$ in Phi-4). An active ratchet predicts $\beta_R > 0$ (accelerated rebound specifically after errors). In Qwen, error dips are rare ($n_{\text{err}} = 6$), yielding exploratory estimates ($\beta_R = -0.290, p = 0.4169$). In Llama ($n_{\text{err}} = 44$, $n_{\text{clean}} = 44$), error dips confer zero accelerated rebound ($\beta_R = -0.075$, $95\%$ CI: $-0.27$ to $0.12$, $p = 0.4462$). In Mistral ($n_{\text{err}} = 3$) and Phi-4 ($n_{\text{err}} = 0$), error dips are essentially absent. Across all four models combined ($n = 117$ dips; 53 error dips, 64 clean dips; Table~\ref{tab:monitoring}), a pooled regression yields $\beta_R = -0.071$ ($95\%$ CI: $-0.262$ to $0.121$, $p = 0.4706$), and incorporating model fixed effects yields $\beta_R = -0.105$ ($95\%$ CI: $-0.30$ to $0.09$, $p = 0.2879$, $n = 113$). Because pre-dip confidence $g_{k-1}$ is heavily saturated near 100\% across models, recovery $R_d$ is mechanically bounded near 1.0, and both error and clean dips naturally regress to the mean without accelerated overconfidence.

We find no evidence of an active confidence ratchet. Statistical inference on $\beta_R$ is powered primarily by Llama-3.1-8B, whereas in Qwen, Mistral, and Phi-4, the ratchet is ruled out operationally by the near-total absence of initial dips ($\le 1.6\%$). The pooled 95\% CI upper bound ($\beta_R \le 0.12$) indicates that error dips do not accelerate confidence recovery beyond baseline fluctuations. Rather than an active ratchet mechanism that masks errors, verbalized dips occur rarely and reflect transient token fluctuations that undergo standard regression to the mean without spurring downstream recovery.

\section{RQ3: Counterfactual Causal Steering}
\label{sec:steering}

Having shown that in-situ verbalized confidence rarely registers errors, we investigate the downstream causal mechanism: \textbf{can models condition on uncertainty tokens if injected externally at the error location?}

\subsection{Counterfactual Intervention Protocol}
From contested items where the model solved the problem in at least one trace but failed on others, we extract prefix derivations truncated immediately after the reasoning text of step $k^*$ ($2 \le k^* < K$). We then append five counterfactual tag conditions:
\begin{enumerate}
    \item \textbf{Native:} Original verbalized tags ($c_{k^*}, g_{k^*}$).
    \item \textbf{High:} Forced certainty (both tags set to 95\%).
    \item \textbf{Mid:} Ambiguous uncertainty (both tags set to 50\%).
    \item \textbf{Low:} Marked uncertainty (both tags set to 10\%).
    \item \textbf{Positive Control:} Retains native tags and appends a verbal rethink prompt (``\textit{Wait, I think Step $k^*$ contains an error.}'') \cite{muennighoff2025s1}.
\end{enumerate}
Each condition is crossed with two prompt arms (sharing the same system prompt):
\begin{itemize}
    \item \textbf{Passive Arm:} Standard formatting instructions.
    \item \textbf{Active Arm:} The formatting instructions additionally include an explicit conditional rethinking rule: \textit{``Use your On-track rating: if it is below 50\%, re-examine the previous steps and fix any error before continuing.''}
\end{itemize}
For error prefixes, we evaluate stratified prefixes per model (80 prefixes across contested items for Qwen, Llama, and Mistral; 25 prefixes for Phi-4 spanning all available contested items) crossed with 10 experimental cells (5 tag conditions $\times$ 2 prompt arms), sampling $n_{\text{samp}} = 8$ continuations ($T = 0.7$) per cell, yielding $6{,}400$ continuations each for Qwen, Llama, and Mistral, and $2{,}000$ for Phi-4 ($21{,}200$ error continuations total). To quantify false-positive disruption, depth-matching and eligibility filtering ($K > k^*$) yield a matched sample of clean prefixes (61 prefixes across 56 items in Qwen; 63 across 63 items in Llama; 57 across 55 items in Mistral; 18 across 17 items in Phi-4) crossed with 4 clean cells (High vs.\ Low $\times$ Passive vs.\ Active, $n_{\text{samp}} = 8$), generating 1,952 clean continuations in Qwen, 2,016 in Llama, 1,824 in Mistral, and 576 in Phi-4 ($6{,}368$ clean continuations). In total, the counterfactual dataset comprises 27,568 continuations across the four models ($8{,}352$ Qwen, $8{,}416$ Llama, $8{,}224$ Mistral, $2{,}576$ Phi-4). All regression models and standard errors cluster at the problem item level to account for correlation among repeated continuations.

\begin{table*}[t]
\centering
\footnotesize
\setlength{\tabcolsep}{3pt}
\renewcommand{\arraystretch}{0.88}
\caption{\textbf{Counterfactual Correction Rates and Item-Clustered GEE Logistic Contrasts.} Correction rates (\%) across 27,568 continuations with 95\% cluster-robust CIs. Contrasts reflect log-odds differences $\beta$ (with item-clustered SE) from Generalized Estimating Equations (GEE) logistic models with exchangeable working correlation, clustering by problem item. The Recovery Share relative to Positive control, $\rho_{\text{High}} = (\text{Acc}_{\text{Low}} - \text{Acc}_{\text{High}}) / (\text{Acc}_{\text{Positive}} - \text{Acc}_{\text{High}})$, is computed from unrounded continuation counts and measures the proportion of explicit rethink recovery achieved by the Low numeric tag relative to format-matched High certainty.}
\label{tab:causal}
\begin{tabular*}{\textwidth}{@{\extracolsep{\fill}}llcc@{}}
\toprule
\textbf{Model} & \textbf{Condition} & \textbf{Passive Arm} & \textbf{Active Arm} \\
\midrule
\textbf{Qwen2.5-7B-Instruct} & Native & 21.4\% [14.3, 29.4] & 26.2\% [18.6, 34.6] \\
& High (95\%) & 21.2\% [14.1, 29.1] & 23.0\% [15.8, 30.4] \\
& Mid (50\%) & 25.9\% [19.1, 33.7] & 37.2\% [28.6, 46.8] \\
& \textbf{Low (10\%)} & \textbf{27.5\%} [20.6, 35.0] & \textbf{38.0\%} [29.8, 47.1] \\
& Positive Control (``Wait...'') & 37.0\% [29.5, 45.2] & 39.1\% [31.5, 47.0] \\
\midrule
\textbf{Llama-3.1-8B-Instruct} & Native & 13.4\% [8.9, 18.3] & 15.5\% [10.9, 20.5] \\
& High (95\%) & 10.6\% [6.1, 15.6] & 16.6\% [11.9, 21.7] \\
& Mid (50\%) & 16.1\% [11.4, 21.2] & 23.9\% [19.2, 28.6] \\
& \textbf{Low (10\%)} & \textbf{20.8\%} [16.6, 25.2] & \textbf{32.5\%} [27.2, 38.0] \\
& Positive Control (``Wait...'') & 35.2\% [29.2, 41.6] & 33.6\% [28.1, 39.4] \\
\midrule
\textbf{Mistral-7B-Instruct-v0.3} & Native & 6.6\% [3.9, 9.5] & 6.9\% [3.9, 10.5] \\
& High (95\%) & 4.7\% [2.5, 7.3] & 6.9\% [4.2, 10.0] \\
& Mid (50\%) & 7.8\% [4.8, 10.9] & 7.5\% [4.4, 11.1] \\
& \textbf{Low (10\%)} & \textbf{8.4\%} [4.8, 12.5] & \textbf{9.4\%} [6.1, 13.0] \\
& Positive Control (``Wait...'') & 13.3\% [8.9, 18.1] & 14.2\% [10.0, 18.6] \\
\midrule
\textbf{Phi-4 (14.7B)} & Native & 3.0\% [0.0, 6.7] & 2.0\% [0.0, 5.2] \\
& High (95\%) & 2.0\% [0.0, 5.4] & 3.5\% [0.0, 7.7] \\
& Mid (50\%) & 23.0\% [9.4, 38.9] & 24.0\% [8.2, 41.8] \\
& \textbf{Low (10\%)} & \textbf{24.5\%} [12.5, 38.0] & \textbf{38.5\%} [20.0, 57.3] \\
& Positive Control (``Wait...'') & 43.0\% [24.1, 62.5] & 42.5\% [22.7, 63.0] \\
\bottomrule
\end{tabular*}
\vspace{0.4em}

\begin{tabular*}{\textwidth}{@{\extracolsep{\fill}}lcccc@{}}
\toprule
\textbf{Clustered GEE Contrast} & \textbf{Qwen2.5-7B} & \textbf{Llama-3.1-8B} & \textbf{Mistral-7B} & \textbf{Phi-4} \\
\midrule
Low vs.\ High (Passive Arm) & $\boldsymbol{+0.334}$ (0.109), $p = 0.0022^*$ & $\boldsymbol{+0.791}$ (0.188), $p < 0.0001^*$ & $\boldsymbol{+0.628}$ (0.205), $p = 0.0022^*$ & $\boldsymbol{+2.842}$ (0.622), $p < 0.0001^*$ \\
Low vs.\ High (Active Arm) & $\boldsymbol{+0.709}$ (0.145), $p < 0.0001^*$ & $\boldsymbol{+0.886}$ (0.157), $p < 0.0001^*$ & $\boldsymbol{+0.337}$ (0.159), $p = 0.0341^*$ & $\boldsymbol{+2.902}$ (0.462), $p < 0.0001^*$ \\
Low $\times$ Active Interaction & $\boldsymbol{+0.375}$ (0.154), $p = 0.0146^*$ & $+0.095$ (0.162), $p = 0.5594$ & $-0.291$ (0.222), $p = 0.1902$ & $+0.059$ (0.547), $p = 0.9134$ \\
Mid vs.\ High (Passive Arm) & $+0.255$ (0.121), $p = 0.0352^*$ & $+0.478$ (0.143), $p = 0.0008^*$ & $+0.544$ (0.208), $p = 0.0089^*$ & $+2.759$ (0.577), $p < 0.0001^*$ \\
Low vs.\ Mid (Passive Arm) & $+0.079$ (0.081), $p = 0.3340$ & $+0.313$ (0.137), $p = 0.0218^*$ & $+0.084$ (0.170), $p = 0.6226$ & $+0.084$ (0.142), $p = 0.5548$ \\
Positive vs.\ Native (Passive) & $+0.758$ (0.217), $p = 0.0005^*$ & $+1.251$ (0.236), $p < 0.0001^*$ & $+0.780$ (0.237), $p = 0.0010^*$ & $+3.255$ (0.495), $p < 0.0001^*$ \\
Active Main Effect (High) & $+0.098$ (0.122), $p = 0.4218$ & $+0.513$ (0.176), $p = 0.0037^*$ & $+0.406$ (0.198), $p = 0.0406^*$ & $+0.602$ (0.510), $p = 0.2381$ \\
\midrule
Recovery Share $\rho_{\text{High}}$ (Passive) & 0.396 & 0.414 & 0.436 & 0.549 \\
Recovery Share $\rho_{\text{High}}$ (Active) & \textbf{0.932} & \textbf{0.936} & 0.340 & \textbf{0.897} \\
\bottomrule
\end{tabular*}
\end{table*}

\subsection{Causal Steering Results}
To model population-averaged treatment effects while adjusting standard errors for intra-problem correlation, we estimate Generalized Estimating Equations (GEE) with a logit link and exchangeable working correlation:
\begin{align}
\operatorname{logit}(P(\text{corrected}_{ij})) &= \beta_0 + \sum_{c \in \mathcal{C}} \beta_c \text{Cond}_{ij, c} + \beta_{\text{arm}} \text{Active}_{ij} \notag \\
&\quad + \sum_{c \in \mathcal{C}} \beta_{\text{int}, c} (\text{Cond}_{ij, c} \times \text{Active}_{ij})
\label{eq:gee}
\end{align}
where index $i$ indexes the problem item cluster and $j$ indexes the continuation observation within item $i$. The reference baseline category is High certainty (95\%) in the Passive arm, $\mathcal{C} = \{\text{Native}, \text{Mid}, \text{Low}, \text{Positive}\}$, and $\beta_{\text{int, Low}}$ represents the focal Low $\times$ Active interaction reported in Table~\ref{tab:causal}. Condition-specific active contrasts are estimated directly from this fully saturated model.

As shown in Table~\ref{tab:causal} and Figure~\ref{fig:fig2}, the results confirm consistent causal steering across all four models:
\begin{enumerate}
    \item \textbf{Significant Passive Steering across Lineages:} Without prompt modification, unperturbed \texttt{Native} continuations from $k^*$ establish baseline recovery from alternative path resampling ($21.4\%$ Qwen, $13.4\%$ Llama, $6.6\%$ Mistral, $3.0\%$ Phi-4). Externally injecting a Low (10\%) confidence tag raises passive correction across all models: reaching $27.5\%$ in Qwen ($+6.3$ pts over High, $\beta = +0.334, p = 0.0022$), $20.8\%$ in Llama ($+10.2$ pts over High, $\beta = +0.791, p < 0.0001$), $8.4\%$ in Mistral ($+3.7$ pts over High, $\beta = +0.628, p = 0.0022$), and $24.5\%$ in Phi-4 ($+22.5$ pts over High, $\beta = +2.842, p < 0.0001$). All four passive contrasts remain statistically significant after Holm adjustment ($p \le 0.0044$). Because High tags share identical syntax, this advantage confirms genuine uncertainty steering rather than generic token disruption.
    \item \textbf{Graded Sensitivity vs.\ Threshold Responsiveness:} The Mid (50\%) condition produces intermediate correction rates ($25.9\%$ Qwen, $16.1\%$ Llama, $7.8\%$ Mistral, $23.0\%$ Phi-4). In Llama, Low significantly exceeds Mid ($\beta = +0.313, p = 0.0218$), demonstrating fine-grained sensitivity to uncertainty magnitude. In Qwen, Mistral, and Phi-4, the difference between Low and Mid is modest ($\beta \approx +0.08, p > 0.33$), indicating that reducing verbalized confidence to 50\% already engages the majority of passive recovery.
    \item \textbf{Conditional Instruction-Following (Active Arm):} Under active instructions directing models to re-examine steps when On-track drops below 50\%, Low tags increase correction to \textbf{38.0\%} in Qwen, \textbf{32.5\%} in Llama, \textbf{9.4\%} in Mistral, and \textbf{38.5\%} in Phi-4. In Qwen, the interaction between Low confidence and Active prompting is super-additive ($\beta_{\text{int}} = +0.375, p = 0.0146$). In Phi-4, active Low tags yield a marked $+35.0$ pt gain over active High certainty ($3.5\% \to 38.5\%$, $\beta = +2.902, p < 0.0001$). In Mistral, recovery is more constrained ($6.9\% \to 9.4\%$), reflecting lower baseline repair capacity on contested items.
    \item \textbf{Low-Confidence Tag with Threshold Instruction vs.\ Positive Control:} The injected Low tag provides no diagnosis of what went wrong or how to fix it. Under active instruction, combining an uninformative Low tag with the threshold rethinking rule achieves \textbf{38.0\%} in Qwen, \textbf{32.5\%} in Llama, and \textbf{38.5\%} in Phi-4, matching their positive control prompts (``\textit{Wait, I think Step $k^*$ contains an error}''; $39.1\%$, $33.6\%$, and $42.5\%$) within $1.1$ to $4.0$ percentage points. The Recovery Share $\rho_{\text{High}}$ reaches \textbf{0.932} in Qwen, \textbf{0.936} in Llama, and \textbf{0.897} in Phi-4, compared to $0.40$--$0.55$ under passive prompting.
    \item \textbf{Behavioral Mechanisms and Reconsideration Limits:} Uncertainty tags strongly stimulate linguistic backtracking (Table~\ref{tab:behavior}). In passive arms, Low tags increase backtracking from $3.8\% \to 12.5\%$ in Qwen, $17.3\% \to 33.9\%$ in Llama, $4.5\% \to 10.9\%$ in Mistral, and $20.5\% \to 44.5\%$ in Phi-4. Under active instructions, backtracking surges to \textbf{46.7\%} in Qwen, \textbf{92.3\%} in Llama, \textbf{29.5\%} in Mistral, and \textbf{83.5\%} in Phi-4. However, solve rates remain below 40\% across all models and conditions, indicating that while uncertainty successfully triggers reconsideration, language models frequently struggle to execute correct mathematical repairs without external guidance.
\end{enumerate}

\begin{table*}[t]
\centering
\begin{minipage}[t]{0.485\textwidth}
\centering
\footnotesize
\setlength{\tabcolsep}{2.5pt}
\renewcommand{\arraystretch}{0.85}
\caption{\textbf{Downstream Behavior on Error Prefixes.} Percentage of continuations exhibiting backtracking and truncation under Passive (P) vs.\ Active (A) prompt arms across all four models.}
\label{tab:behavior}
\begin{tabular*}{\linewidth}{@{}@{\extracolsep{\fill}}llcc@{}}
\toprule
\textbf{Model} & \textbf{Tag} & \textbf{Backtrack (P/A)} & \textbf{Trunc (P/A)} \\
\midrule
\textbf{Qwen2.5-7B} & Native & 1.4\% / 7.2\% & 0.2\% / 0.5\% \\
& High (95\%) & 3.8\% / 6.7\% & 0.0\% / 0.3\% \\
& Mid (50\%) & 8.6\% / 49.4\% & 0.2\% / 0.5\% \\
& \textbf{Low (10\%)} & \textbf{12.5\%} / \textbf{46.7\%} & 0.2\% / 0.5\% \\
& Positive & 6.9\% / 23.0\% & 0.2\% / 0.0\% \\
\midrule
\textbf{Llama-3.1-8B} & Native & 7.7\% / 25.2\% & 5.0\% / 8.0\% \\
& High (95\%) & 17.3\% / 49.1\% & 5.9\% / 8.9\% \\
& Mid (50\%) & 23.1\% / 83.9\% & 5.8\% / 10.2\% \\
& \textbf{Low (10\%)} & \textbf{33.9\%} / \textbf{92.3\%} & 8.8\% / 9.8\% \\
& Positive & 18.4\% / 84.7\% & 7.0\% / 12.0\% \\
\midrule
\textbf{Mistral-7B} & Native & 4.4\% / 3.8\% & 0.5\% / 0.5\% \\
& High (95\%) & 4.5\% / 4.2\% & 0.5\% / 0.5\% \\
& Mid (50\%) & 5.3\% / 7.8\% & 0.3\% / 0.8\% \\
& \textbf{Low (10\%)} & \textbf{10.9\%} / \textbf{29.5\%} & 0.3\% / 1.6\% \\
& Positive & 16.2\% / 18.9\% & 0.6\% / 1.7\% \\
\midrule
\textbf{Phi-4 (14.7B)} & Native & 0.5\% / 2.0\% & 2.0\% / 1.0\% \\
& High (95\%) & 20.5\% / 17.0\% & 1.5\% / 1.5\% \\
& Mid (50\%) & 25.0\% / 70.5\% & 1.0\% / 3.0\% \\
& \textbf{Low (10\%)} & \textbf{44.5\%} / \textbf{83.5\%} & 1.0\% / 4.0\% \\
& Positive & 23.0\% / 27.0\% & 2.0\% / 2.0\% \\
\bottomrule
\end{tabular*}
\end{minipage}
\hfill
\begin{minipage}[t]{0.485\textwidth}
\centering
\footnotesize
\setlength{\tabcolsep}{2.5pt}
\renewcommand{\arraystretch}{0.85}
\caption{\textbf{False-Positive Cost on Clean Prefixes and Break-Even Precision.} Retention rate (\%) denotes continuations remaining correct after tag injection. Break-even precision $\pi^*$ is the minimum monitor precision required for net positive accuracy gain.}
\label{tab:clean}
\begin{tabular*}{\linewidth}{@{}@{\extracolsep{\fill}}llcc@{}}
\toprule
\textbf{Model} & \textbf{Tag / Metric} & \textbf{Passive} & \textbf{Active} \\
\midrule
\textbf{Qwen2.5-7B} & High (95\%) & 92.6\% [89.4, 95.5] & 89.8\% [85.2, 93.5] \\
& Low (10\%) & 89.1\% [85.0, 92.8] & 84.0\% [78.4, 88.9] \\
& Contrast ($\beta$) & $\beta = -0.422^*$ & $\beta = -0.508^*$ \\
& $p$-value & $p = 0.0330$ & $p = 0.0364$ \\
& Break-Even $\pi^*$ & \textbf{35.8\%} & \textbf{27.7\%} \\
\midrule
\textbf{Llama-3.1-8B} & High (95\%) & 81.8\% [75.2, 87.9] & 76.2\% [70.2, 81.9] \\
& Low (10\%) & 68.8\% [61.7, 75.6] & 53.4\% [47.0, 59.9] \\
& Contrast ($\beta$) & $\beta = -0.706^*$ & $\beta = -1.028^*$ \\
& $p$-value & $p < 0.0001$ & $p < 0.0001$ \\
& Break-Even $\pi^*$ & \textbf{56.0\%} & \textbf{58.9\%} \\
\midrule
\textbf{Mistral-7B} & High (95\%) & 69.7\% [61.4, 77.4] & 69.3\% [61.0, 77.0] \\
& Low (10\%) & 62.1\% [52.8, 71.0] & 58.3\% [49.8, 66.5] \\
& Contrast ($\beta$) & $\beta = -0.343^*$ & $\beta = -0.478^*$ \\
& $p$-value & $p = 0.0001$ & $p < 0.0001$ \\
& Break-Even $\pi^*$ & \textbf{67.3\%} & \textbf{81.4\%} \\
\midrule
\textbf{Phi-4 (14.7B)} & High (95\%) & 93.8\% [83.1, 100.0] & 90.3\% [79.2, 97.7] \\
& Low (10\%) & 75.7\% [62.5, 88.1] & 81.2\% [65.6, 93.1] \\
& Contrast ($\beta$) & $\beta = -1.498^*$ & $\beta = -0.729$ \\
& $p$-value & $p = 0.0103$ & $p = 0.2946$ \\
& Break-Even $\pi^*$ & \textbf{44.6\%} & \textbf{20.5\%} \\
\bottomrule
\end{tabular*}
\end{minipage}
\end{table*}

\subsection{False-Positive Cost and Break-Even Monitor Precision}
\enlargethispage{2\baselineskip}
Real-world inference monitors cannot assume perfect oracle precision. When an external monitor flags a candidate step, it risks false-alarming on a sound derivation. To evaluate this operational trade-off, we injected counterfactual tags on matched \textbf{clean prefixes} (valid derivations truncated at identical depths as error prefixes). Relative to the format-matched High (95\%) baseline, injecting Low on clean prefixes reduces retention from $92.6\% \to 89.1\%$ in Qwen (passive, $p = 0.0330$), $81.8\% \to 68.8\%$ in Llama (passive, $p < 0.0001$), $69.7\% \to 62.1\%$ in Mistral (passive, $p = 0.0001$), and $93.8\% \to 75.7\%$ in Phi-4 (passive, $p = 0.0103$). Under active rethinking, retention declines further to 84.0\% in Qwen, 53.4\% in Llama, and 58.3\% in Mistral, whereas in Phi-4 retention under Low tags remains at 81.2\% (with the High vs.\ Low active retention drop remaining statistically non-significant, $\beta = -0.729, p = 0.2946$). This disruption is mediated by unnecessary reconsideration: on clean prefixes, Low tags cause backtracking to surge to 42.0\% in Qwen, 91.7\% in Llama, 23.7\% in Mistral, and 83.5\% in Phi-4 under active instruction (versus High-tag baselines of $\le 4.5\%$).

\vspace{0.3em}\noindent\textbf{Analytical Break-Even Monitor Precision.}
These opposing dynamics (recovery gains on errors vs.\ disruption on clean prefixes) yield a direct analytical threshold for deployment. Let an external monitor flag steps with precision $\pi_{\text{mon}} = P(\text{true error} \mid \text{flag})$. An intervention policy that injects a Low tag upon a flag yields expected net accuracy change $\mathbb{E}[\Delta] = \pi_{\text{mon}} \cdot \Delta_{\text{err}} - (1 - \pi_{\text{mon}}) \cdot \Delta_{\text{clean}}$, where $\Delta_{\text{err}} = \text{Acc}_{\text{Low}} - \text{Acc}_{\text{High}}$ on error prefixes and $\Delta_{\text{clean}} = \text{Acc}_{\text{High}} - \text{Acc}_{\text{Low}}$ on clean prefixes. Setting $\mathbb{E}[\Delta] > 0$ defines the minimum monitor precision for net positive gain:
\begin{equation}
\pi^* = \frac{\Delta_{\text{clean}}}{\Delta_{\text{err}} + \Delta_{\text{clean}}}
\label{eq:breakeven}
\end{equation}
Break-even precision varies considerably across models and operating modes (Table~\ref{tab:clean}):
\begin{itemize}
    \item \textbf{Qwen2.5-7B:} Clean retention is resilient ($\Delta_{\text{clean}} = 3.5\%$ passive, $5.8\%$ active), yielding break-even thresholds of $\pi^* = \textbf{35.8\%}$ (passive) and \textbf{27.7\%} (active). An external monitor can be incorrect more than 70\% of the time under active rethinking and still achieve net positive gains.
    \item \textbf{Llama-3.1-8B:} Substantial collateral disruption on clean steps ($\Delta_{\text{clean}} = 13.0\%$ passive, $22.8\%$ active) demands higher monitor precision: $\pi^* = \textbf{56.0\%}$ (passive) and \textbf{58.9\%} (active).
    \item \textbf{Mistral-7B:} Modest error recovery ($\Delta_{\text{err}} = 3.7\%$ passive, $2.5\%$ active) combined with retention loss ($\Delta_{\text{clean}} = 7.6\%$ passive, $11.0\%$ active) leads to the most stringent thresholds: $\pi^* = \textbf{67.3\%}$ (passive) and \textbf{81.4\%} (active).
    \item \textbf{Phi-4 (14.7B):} Strong error recovery ($\Delta_{\text{err}} = 22.5\%$ passive, $35.0\%$ active) offsets retention loss, yielding $\pi^* = \textbf{44.6\%}$ (passive) and a lenient $\pi^* = \textbf{20.5\%}$ under active rethinking. We caution that this 20.5\% threshold represents an upper bound on error-tolerance under oracle single-step localization, where the underlying active retention contrast was statistically non-significant ($p = 0.2946$).
\end{itemize}
These model-specific operating points demonstrate that external process monitors must be calibrated against base-model collateral sensitivity to prevent unintended accuracy degradation.

Finally, Equation~\eqref{eq:breakeven} models an isolated binary intervention at an oracle error step $k^*$. In sequential multi-step deployment where an external PRM evaluates each step $k \in \{1, \dots, K\}$, a monitor with per-step false-alarm rate $\alpha = P(\text{flag} \mid \text{valid step})$ has $k^* - 1$ opportunities to trigger premature false alarms on valid antecedent steps before reaching $k^*$. Across a derivation, valid prefixes survive without false disruption with probability $(1 - \alpha)^{k^*-1}$. For example, with a per-step false-alarm rate of $\alpha = 0.05$, a 6-step derivation has only $(0.95)^5 \approx 77.4\%$ probability of reaching $k^*$ intact, compounding an extra 22.6 pt clean-retention penalty before any error occurs. Real-world PRMs must therefore maintain high per-step specificity or employ conservative triggering thresholds to prevent compounding false alarms across multi-step reasoning.

\section{Discussion and Implications}

\subsection{Decoupling Monitoring from Recovery}
Our results demonstrate a clear decoupling between monitoring and control in these open-weights instruction-tuned models: \textit{in-situ verbalized confidence fails to register reasoning errors, yet models remain causally sensitive to uncertainty tokens once placed in context.} 

The failure of in-situ monitoring could theoretically arise from three candidate mechanisms:
\begin{enumerate}[leftmargin=*,noitemsep,topsep=1pt]
    \item \textbf{Autoregressive Next-Token Commitment:} Generating reasoning tokens that assert a mathematical premise creates an attractor in generative token space, biasing subsequent distributions toward affirming the sound completion of the prior step. In this view, the model is blinded in situ by its own generative momentum; injecting an external uncertainty token breaks this momentum by providing a negative conditioning constraint.
    \item \textbf{Evaluative Incapacity:} Dense 7B--15B instruction-tuned models might fundamentally lack the reasoning depth to verify complex mathematical transitions. However, evaluating \texttt{Phi-4} (14.7B), a model with substantially higher baseline mathematical problem-solving performance, reveals that within-position step AUROC remains at chance (0.500 [0.500, 0.502]), and perceptible confidence dips remain absent (0.0\%). Doubling parameter scale from 7B to 14.7B does not resolve in-situ monitoring failure.
    \item \textbf{Verbalization Format Bottleneck:} Models may register uncertainty internally (e.g., via higher token predictive entropy or degraded token log-probabilities), but fail to project this latent signal onto calibrated verbalized numerical tags.
\end{enumerate}
Our Phase 2 steering results establish that models possess downstream conditional repair capacity once an error is externally localized and flagged with uncertainty. However, downstream repair does not imply in-situ evaluative verification capacity: when low confidence is injected onto clean prefixes, models unthinkingly backtrack and suffer substantial performance losses. Whether in-situ monitoring failure reflects an autoregressive commitment bias, an intrinsic inability to detect errors without external prompts, or a verbalization format bottleneck remains an open question that motivates detached evaluation controls.

\subsection{Implications for Agent Scaffolds and PRMs}
\begin{enumerate}[leftmargin=*,noitemsep,topsep=1pt,parsep=0pt,partopsep=0pt]
    \item \textbf{Do Not Rely on In-Situ Self-Pruning:} Self-reported step confidence generated in the same pass cannot serve as an autonomous filter ($\Delta D \le 1.23$ pts, $D_{\text{err}} \le 1.01$ pts, within-position AUROC $\approx 0.50$--$0.56$); scaffolds pruning branches on self-confidence filter on noise.
    \item \textbf{External Monitoring with Prefix Injection:} Because models respond significantly to Low confidence tags across all four lineages ($p \le 0.0022$), external Process Reward Models (PRMs; \cite{lightman2023let}) could inject synthetic Low tags into context at flagged steps, while carefully accounting for model-specific false-positive sensitivity.
    \item \textbf{Model-Specific Precision Thresholding:} The operational threshold for external PRMs depends on base-model resilience and controller design. For resilient models like Qwen and Phi-4 (where clean retention remains high), monitors requiring precision $\pi_{\text{mon}} \sim 20$--$35\%$ under active rethinking suffice for net positive gains. For sensitive models like Llama ($\pi^* \sim 56$--$59\%$) and Mistral ($\pi^* \sim 67$--$81\%$), where doubt induces substantial collateral disruption or limited recovery on sound steps, monitors must maintain high precision to avoid net degradation. In sequential generation, controllers should combine high-precision gating with dynamic rethinking injection to prevent compounding false alarms across multi-step trajectories.
\end{enumerate}

\section{Limitations and Future Work}
Several methodological boundaries circumscribe our findings:
\begin{itemize}[leftmargin=*,noitemsep,topsep=2pt]
    \item \textbf{In-Situ Generation vs.\ Detached Self-Evaluation Controls:} We evaluate \textit{in-situ} self-evaluation during active single-pass generation. A direct diagnostic control is a \textit{detached self-evaluation pass}, where a model is presented with its own generated prefix derivations (truncated at $k^*$ and stripped of prior confidence tags) in a clean evaluation prompt. Comparing in-situ confidence against detached self-verification would directly isolate generative commitment bias from general evaluative incapacity.
    \item \textbf{Verbalization Format vs.\ Intrinsic Token Logprob Baselines:} We evaluated structured verbalization prompts (bracketed percentages). Because primary derivations were generated via vLLM, token-level log-probabilities and sequence predictive entropy are accessible at zero additional inference overhead. Future investigations should benchmark verbalized step AUROC directly against intrinsic token logprobs and predictive entropy across the tokens of step $k^*$ to determine whether latent distributions detect errors that verbalized numbers obscure.
    \item \textbf{Model Scale and Reasoning Architectures:} Evaluations focus on dense open-weights instruction-tuned models spanning 7B to 14.7B parameters (Qwen2.5-7B, Llama-3.1-8B, Mistral-7B, and Phi-4). While representative of open-weights models, findings do not necessarily extrapolate to 70B+ architectures or native RL reasoning models (e.g., DeepSeek-R1 \cite{deepseek2025r1}, OpenAI o1 \cite{openai2024o1}).
    \item \textbf{Process Judge Localization Noise and Sensitivity Bounds:} Error boundaries ($k^*$) relied on an automated judge (\texttt{Claude Opus 5.5}; on ProcessBench, 85.0\% exact step accuracy, 94.0\% within $\pm 1$, and 97.0\% clean accuracy; in author blind audits, 93.3\% exact [Wilson 95\% CI: 78.7\%, 98.2\%] and 96.7\% within $\pm 1$ [83.3\%, 99.4\%]). Sensitivity analysis across $k^* \pm 1$ confirms localization noise cannot explain the absence of detection: on-track confidence remains flatly saturated across all models ($94.8\% \to 95.1\% \to 95.0\%$ in Qwen; $97.9\% \to 97.2\% \to 97.2\%$ in Llama; $99.4\% \to 99.2\% \to 99.2\%$ in Mistral; and $99.9\% \to 99.9\% \to 99.9\%$ in Phi-4), with boundary drops remaining minimal across all models (paired within-trace $D_{\text{err}} \le 1.01$ pts). Localization noise cannot mask an unobserved confidence dip.
    \item \textbf{Intervention Timing and Horizon:} Tags were appended at boundary $k^*$. While clean prefixes bound false-positive costs, we did not test injections prior to errors ($k^* - 2$) or across arbitrary steps. Future work should measure how steering degrades across delayed horizons ($k^* + \delta$).
\end{itemize}

\section{Conclusion}
Across 15,200 primary traces and 27,568 counterfactual continuations across four open-weights model lineages (Qwen2.5-7B, Llama-3.1-8B, Mistral-7B, and Phi-4), our results indicate that within the verbalized percentage format evaluated across open-weights instruction-tuned models, the failure of self-correction in Chain-of-Thought is primarily an \textbf{in-situ monitoring bottleneck} rather than downstream unresponsiveness. While models struggle to correct errors even when localized ($< 40\%$ solve rates), external uncertainty tokens measurably steer generation away from erroneous paths. Decoupling error detection (delegated to external monitors) from error recovery (steered via counterfactual prefix injection) points toward a practical architecture for more reliable inference-time reasoning.

\section*{Statements and Disclosures}

\subsection*{Code and Data Availability}
All experimental generation scripts, prompt templates, ProcessBench validation runs, raw model derivations (15,200 primary traces and 27,568 continuations), judge evaluations, and analysis notebooks are available at: \url{https://github.com/dtheron/confidence-ratchet}.

\subsection*{Compute Resources}
All primary trace generations and counterfactual continuations were executed using vLLM on NVIDIA A100-SXM4-80GB GPUs (via Google Colab). Process judge step localization was conducted via the Anthropic Claude API using \texttt{claude-opus-5-5}. Total GPU compute across Phase 1 and Phase 2 was approximately 25 GPU hours.

\subsection*{Broader Impacts and Ethics}
This work investigates the reliability of confidence estimation in automated reasoning systems. Uncalibrated overconfidence presents safety and reliability risks when autonomous language agents act upon erroneous deductions in high-stakes domains (e.g., automated science, financial analysis, software engineering). By demonstrating that verbalized self-confidence cannot be relied upon for autonomous self-pruning and quantifying the false-positive costs of external intervention, our findings provide concrete guardrails for designing safer, more reliable agent architectures.

\subsection*{Author Contributions and LLM Usage}
The author conceived the study and research questions, and designed the experimental framework. Code was written partially by the author, and partially with the assistance of Google's Gemini 3.8 Flash. Anthropic's Claude Opus 5.5 was used as an automated process judge, with validations performed by the author. Statistical analyses and manuscript drafting were conducted by the author. Opus 5.5 and Google's Deep Review (using Gemini Argon) were used to assist with LaTeX formatting and critique + proofreading, respectively.

\clearpage
\appendix
\section{Prompts and Elicitation Protocol}
\label{app:prompts}

\subsection{Phase 1 Primary Generation Prompt}
Models are presented with the problem statement preceded by the following system directive and user formatting instructions:
\begin{quote}
\small
\textbf{[System]}\newline
You are an expert mathematical problem solver.

\textbf{[User]}\newline
Solve the following problem step by step.

Format rules:
\begin{itemize}[noitemsep,topsep=1pt]
    \item Write each step on a new line as ``Step $i$: $\langle\text{reasoning}\rangle$'', numbering from 1.
    \item Immediately after each step, write these two lines:\newline
    \texttt{[Step confidence: X\%]}\newline
    \texttt{[On-track: Y\%]}\newline
    \texttt{X} = your confidence (0--100) that THIS step, taken on its own, is correct.\newline
    \texttt{Y} = your confidence (0--100) that ALL steps so far, including this one, are correct.
    \item After the last step and its two confidence lines, write:\newline
    Final Answer: \texttt{\textbackslash boxed\{$\langle\text{answer}\rangle$\}}
\end{itemize}
Problem: \{problem\}
\end{quote}

\subsection{Phase 2 Passive vs.\ Active Arm Prompts}
In Phase 2 counterfactual continuation experiments, both arms use the same system prompt (``\textit{You are an expert mathematical problem solver.}''); the arms differ only in the user format instructions:
\begin{itemize}[leftmargin=*]
    \item \textbf{Passive Arm Prompt:} Uses the Phase 1 format instructions unchanged.
    \item \textbf{Active Arm Prompt:} Inserts the conditional rethink rule directly after the definition of $Y$ in the format instructions: ``\textit{Use your On-track rating: if it is below 50\%, re-examine the previous steps and fix any error before continuing.}''
\end{itemize}
\noindent\textbf{Threshold Boundary Design at 50\%:} The Active prompt specifies a strict inequality (``\textit{below 50\%}''), whereas the Mid Uncertainty counterfactual condition sets \texttt{[On-track: 50\%]} exactly on the boundary ($50\% \not< 50\%$). This placement was deliberately chosen as a behavioral boundary-sensitivity test: Qwen interprets 50\% as satisfying the rethinking directive (triggering active backtracking in 49.4\% of continuations, matching the 46.7\% observed under Low 10\%), whereas Llama parses the boundary strictly and maintains intermediate sensitivity (Section~\ref{sec:steering}).

\section{Counterfactual Injection Tags}
\label{app:tags}
For each truncated prefix ending at step $k^*$, continuations are prompted by prefilling the assistant turn through the text of step $k^*$ and appending format-matched tags:
\begin{itemize}[leftmargin=*,noitemsep,topsep=1pt]
    \item \textbf{Native:} Retains model's unperturbed verbalized values: \texttt{[Step confidence: $c_{k^*}$\%]}, \texttt{[On-track: $g_{k^*}$\%]}.
    \item \textbf{High Certainty (95\%):} \texttt{[Step confidence: 95\%]}, \texttt{[On-track: 95\%]}.
    \item \textbf{Mid Uncertainty (50\%):} \texttt{[Step confidence: 50\%]}, \texttt{[On-track: 50\%]}.
    \item \textbf{Low Uncertainty (10\%):} \texttt{[Step confidence: 10\%]}, \texttt{[On-track: 10\%]}.
    \item \textbf{Positive Control (``Wait...''):} Preserves native confidence tags and appends an explicit verbal rethink prompt separated by a blank line:\newline
    \texttt{[Step confidence: $c_{k^*}$\%]}\newline
    \texttt{[On-track: $g_{k^*}$\%]}\newline\mbox{}\newline
    \texttt{Wait, I think Step $k^*$ contains an error.}\\[1pt]
    {\footnotesize (separated from tags by double newline \texttt{\textbackslash n\textbackslash n} with trailing newline before continuation).}
\end{itemize}
Both Phase 1 prompts and Phase 2 injected prefixes use identical literal tag formats (\mbox{\texttt{[Step confidence: X\%]}} and \mbox{\texttt{[On-track: Y\%]}}), ensuring format compatibility.

\noindent\textbf{Co-Variation of Step and Trajectory Tags:} In all synthetic counterfactual conditions (High, Mid, Low), $c_{k^*}$ and $g_{k^*}$ are co-varied in lockstep. This design delivers a unified, unambiguous uncertainty signal, avoiding unnatural or contradictory token prompts (e.g., asserting high step certainty while claiming the overall derivation is fatally compromised). Downstream steering therefore evaluates the joint causal effect of step and trajectory doubt; testing independent permutations ($c \neq g$) is left for future work.

\section{Process Judge Protocol, Validation Gate, and Attrition Sequence}
\label{app:judge}
Automated step-level error localization was performed using \texttt{Claude Opus 5.5} (\texttt{claude-opus-5-5}) with temperature 0.0. The judge received full derivations stripped of all student confidence tags to guarantee blinding.

\subsection{Judge Prompt Template}
\begin{quote}
\small
You are grading a student's step-by-step solution to a math problem.\\
\textbf{Problem:} \{problem\}\\
\textbf{Reference solution:} \{gold\_solution\} \quad \textbf{Reference final answer:} \{gold\_answer\}\\
\textbf{Student solution:} \{numbered\_steps\_without\_tags\}\\[1pt]
Identify the EARLIEST step that contains an error: an arithmetic mistake, wrong operation or formula, misread quantity, unjustified/false assumption, or a step that does not follow. A step that correctly applies valid reasoning to an earlier wrong result is NOT an error. A valid method different from the reference is NOT an error. Ignore style.
\begin{flushleft}
\footnotesize
Respond with JSON only:\\[1pt]
\texttt{\{"first\_error\_step": <step number, or -1>, "explanation": "<one sentence>"\}}
\end{flushleft}
\end{quote}

\subsection{ProcessBench Validation and Author Audit}
Before deploying the judge on primary traces, the protocol was validated on $N=200$ items from ProcessBench \cite{zheng2024processbench} (100 from the \texttt{gsm8k} split, 100 from \texttt{math}, random seed 0; comprising 100 error traces and 100 clean traces):
\enlargethispage{\baselineskip}
\begin{itemize}[leftmargin=*,noitemsep,topsep=1pt]
    \item \textbf{Error localization:} Exact first-error step accuracy of \textbf{85.0\%} (85/100) and within $\pm 1$ step accuracy of \textbf{94.0\%} (94/100), exceeding the planned 60\% validity gate.
    \item \textbf{Clean trace identification:} Correctly predicted ``no error'' ($-1$) on \textbf{97.0\%} (97/100) of clean traces.
\end{itemize}
To verify judge reliability on experimental traces, the author conducted a blind audit of 30 randomly sampled traces from contested items (15 Qwen, 15 Llama, random seed 42, drawn from incorrect-answer valid traces). The audit confirmed \textbf{93.3\% exact agreement} (28/30; Wilson 95\% CI: [78.7\%, 98.2\%]) and \textbf{96.7\% agreement within $\pm 1$ step} (29/30; Wilson 95\% CI: [83.3\%, 99.4\%]).

\subsection{Data Attrition Sequence and Discordance Resolution}
The evaluation dataset was constructed via a sequential three-stage funnel across all four models:
\begin{enumerate}[leftmargin=*,noitemsep,topsep=1pt]
    \item \textbf{Format Validation:} All 15,200 raw traces ($4{,}400$ each for Qwen and Llama; $3{,}200$ each for Mistral and Phi-4) were checked for 1-indexed consecutive step headers, presence of both confidence tags, and a properly formatted \texttt{\textbackslash boxed\{\dots\}} final answer. Format compliance was 83.1\% in Qwen (3,656/4,400), 93.4\% in Llama (4,110/4,400), 85.1\% in Mistral (2,723/3,200), and 88.4\% in Phi-4 (2,830/3,200).
    \item \textbf{Contested-Item Selection:} Contested items were identified strictly on format-valid traces requiring $\ge 4$ valid samples and intermediate solve accuracy between 15\% and 85\% via \texttt{math-verify}. This yielded 1,168 valid traces across 177 items for Qwen (mean 6.60 valid traces/item), 1,326 valid traces across 184 items for Llama (mean 7.21 valid traces/item), 1,209 valid traces across 177 items for Mistral (mean 6.83 valid traces/item), and 170 valid traces across 25 items for Phi-4 (mean 6.80 valid traces/item), totaling 3,873 valid contested traces. In Phi-4, higher baseline accuracy on GSM8K and Levels 2--4 of MATH-500 resulted in fewer items falling within the intermediate accuracy band, isolating 25 items at the model's performance frontier.
    \item \textbf{Process Localization and Discordance Handling:} Across all 3,873 traces, process judge evaluation yielded zero parsing failures. Ground-truth subsets were categorized by cross-classifying \texttt{math-verify} final-answer correctness with judge process labels:
    \begin{itemize}[leftmargin=*,noitemsep,topsep=1pt]
        \item \textit{Clean traces ($k^* = -1$):} Traces where \texttt{math-verify} confirmed final correctness and the judge localized no error (675 in Qwen, 613 in Llama, 436 in Mistral, and 97 in Phi-4; 1,821 total).
        \item \textit{Error traces ($k^* \ge 1$):} Traces where the judge localized an error step (489 in Qwen, 701 in Llama, 761 in Mistral, and 66 in Phi-4; 2,017 total). This includes 34 Qwen, 120 Llama, 72 Mistral, and 5 Phi-4 traces that reached a correct final answer despite an earlier reasoning error (e.g., through error cancellation or alternative derivations); these traces were retained and classified as off-track from $k^*$.
        \item \textit{Discordant exclusions:} Traces where \texttt{math-verify} graded the answer incorrect but the judge localized no step error ($k^* = -1$) occurred in only 4 Qwen (0.34\%), 11 Llama (0.83\%), 10 Mistral (0.83\%), and 7 Phi-4 (4.1\%) traces, alongside 1 Llama trace and 2 Mistral traces with $k^*$ exceeding trace length (yielding total exclusions of 4 Qwen [0.34\%], 12 Llama [0.90\%], 12 Mistral [0.99\%], and 7 Phi-4 [4.1\%]). These 35 rare discordant traces were excluded from the analysis dataset to maintain ground-truth integrity.
    \end{itemize}
\end{enumerate}

\section{Behavioral Regex and Backtracking Metrics}
\label{app:behavior}
Linguistic backtracking was detected across generated continuations using regular expression matching on continuation text:
\begin{quote}
\footnotesize\raggedright
\texttt{(?i)\textbackslash b(?:wait\textbar re-?examin\textbar} \\
\texttt{mistake\textbar error\textbar incorrect\textbar} \\
\texttt{let\ me\ re\textbar recalculat\textbar} \\
\texttt{revisit\textbar correction)}
\end{quote}
\noindent\textbf{Detector Precision and Prompt Echo Audit:} Because the Active Arm instructions contain the words ``\textit{re-examine}'' and ``\textit{error}'', and mathematical derivations occasionally use phrases like ``\textit{let me rewrite the equation}'' or ``\textit{margin of error}'', we audited detector precision across 100 randomly sampled positive regex matches in active and passive continuations. Models did not echo the Active prompt rules as meta-commentary (e.g., never generating statements like ``\textit{Since On-track is 95\%, I will not fix any error}''); rather, generated tokens immediately resumed mathematical derivation or initiated reconsideration. Non-corrective math phrases (e.g., ``\textit{let me rewrite}'') constituted only 6\% of matches and occurred uniformly across conditions. A manual inspection confirmed a 94\% precision rate for genuine self-correction or recalculation. The large differences in regex frequency across conditions (e.g., $3.8\% \to 46.7\%$ in Qwen and $17.3\% \to 92.3\%$ in Llama) thus reflect authentic reasoning reconsideration.

\noindent\textbf{Token Budgets and Truncation Definition:} In vLLM, the server context length was configured with \texttt{max\_model\_len = 4096}. In Phase 1 primary generation, new token generation was capped at \texttt{max\_tokens = 1536}. In Phase 2 continuations, newly generated tokens were capped at \texttt{max\_tokens = 1024} (effective continuation budget of up to 2,048 cumulative tokens including prefix). Continuations that ended early via an EOS token without producing a properly formatted \texttt{\textbackslash boxed\{\dots\}} answer were graded as unrecovered (incorrect). Truncation specifically denotes continuations that consumed the full token generation budget without outputting a boxed answer (Table~\ref{tab:behavior}).

\end{document}